%% file: main.tex
\documentclass[11pt]{article}

\usepackage[margin=1in]{geometry}
\usepackage{times}
\usepackage[T1]{fontenc}
\usepackage{amsmath,amssymb}
\usepackage{graphicx}
\usepackage{booktabs}
\usepackage{natbib}
\usepackage{hyperref}
\usepackage{url}
\usepackage{xcolor}
\usepackage{caption}
\usepackage{subcaption}
\usepackage{multirow}
\usepackage{authblk}

\setcitestyle{numbers,square,comma}
\hypersetup{colorlinks=true, linkcolor=blue!50!black, citecolor=blue!50!black, urlcolor=blue!50!black}

\title{Exact Degeneracy Under Balanced $k$-Shot Sampling: Consequences for
Small-Sample Discriminant Analysis on LLM Embeddings}

\author[1]{Lingxiao Qu\thanks{Corresponding author. Email: \texttt{lxqu1110@outlook.com}}}
\affil[1]{The University of Aizu, Aizuwakamatsu, Fukushima, Japan}

\date{}

\begin{document}
\maketitle

\begin{abstract}
\input{sections/abstract.tex}
\end{abstract}

\section{Introduction}
\label{sec:intro}
\input{sections/introduction.tex}

\section{Related Work}
\label{sec:related}
\input{sections/related_work.tex}

\section{Method}
\label{sec:method}
\input{sections/method.tex}

\section{Experimental Setup}
\label{sec:setup}
\input{sections/setup.tex}

\section{Results}
\label{sec:results}
\input{sections/results.tex}

\section{Guidelines}
\label{sec:guidelines}
\input{sections/guidelines.tex}

\section{Discussion and Limitations}
\label{sec:discussion}
\input{sections/discussion.tex}

\section{Conclusion}
\label{sec:conclusion}
\input{sections/conclusion.tex}

\bibliographystyle{plainnat}
\bibliography{references}

\end{document}

%% file: sections/abstract.tex
Balanced $k$-shot sampling, drawing exactly $k$ labeled examples per
class, is the standard protocol in few-shot evaluation. We show that it
induces an \emph{exact} degeneracy in Kernelized Linear Principal
Component Discriminant Analysis (KLPCDA), a family of seven supervised
dimensionality reduction methods. The within-class operator becomes
exactly a scaled orthogonal projector, so two variants have all signal
eigenvalues equal and a third has a zero operator, and the truncated
pseudo-inverse used to implement them removes exactly the subspace that
contains all between-class information. We verify this to float64
precision on sentence embeddings and on residual-stream activations of a
decoder-only model, and find that the accuracies of the truncated
variants change with the number of BLAS threads, by up to 12 percentage
points (57 for the variant with a zero operator).

Replacing the truncation with a ridge selected within the support set,
we evaluate few-shot text classification on frozen embeddings ($d$ up to
4096) over four datasets, three embedding models, and three trained
baselines. The regularized formulation outperforms the best original
variant in 45 of 48 settings, but in general not a tuned linear probe or
nearest centroid on bidirectional encoders. Its gains are largest on
E5-Mistral, the only decoder-based embedding model, and come from
regularized within-class whitening rather than from KLPCDA: shrinkage
LDA outperforms both baselines in all 16 E5-Mistral settings, and
pre-specified tests show that the effect depends on the embedding model
rather than on its dimension. E5-Mistral's remaining gap is largely an
estimation-efficiency effect: increasing $k$ to 30--50 reduces the
probe's gap to a bidirectional encoder by more than 80\%. We give five
guidelines limited to the conditions tested and release the code and
per-seed results.

%% file: sections/introduction.tex
Small-sample-size (SSS) classification, where the number of labeled
examples $n$ is smaller than the feature dimension $d$, has a long history
in classical discriminant analysis. When $n < d$, the within-class scatter
matrix $S_w$ that Fisher's linear discriminant analysis (LDA) must invert
is rank-deficient (\mbox{$\mathrm{rank}(S_w) \le n - L$} for $L$
classes), so naive inversion is numerically unstable or undefined.
Kernelized Linear Principal Component Discriminant Analysis (KLPCDA)
\citep{qu2026klpcda} is a supervised dimensionality reduction method for
this setting. Its seven variants combine total variance $C$,
between-class scatter $S_b$, and within-class scatter $S_w$ in different
ways, and some of them do not use $S_w$ at all. A separate classifier
(1-nearest-neighbor) is then applied to the projected features.

Frozen embeddings from a modern LLM or sentence encoder, used for
few-shot text classification, fall into the same regime: $k$ labeled
examples per class across $L$ classes gives $n = kL$, while embedding
dimensions range from $d=384$ (small sentence-transformers) to $d=4096$
(billion-parameter decoder-based embedding models), so $n \ll d$ for
realistic budgets ($k \le 10$). Here, ``few-shot'' means a supervised
method trained on these $k$ examples per class on top of a frozen
embedding; in-context learning, which the term often refers to for
language models, is included as a baseline. This paper asks what
small-sample discriminant analysis contributes in this setting, compared
with the simple and strong alternatives a practitioner would otherwise
use (a tuned linear probe, nearest centroid, and trained baselines), and
which of its components is responsible for any benefit.

The first part of the answer does not depend on any dataset. Balanced
$k$-shot sampling, the standard protocol, draws exactly $k$ examples per
class. Under this protocol, the KLPCDA family is in an exactly
characterizable degenerate state: the within-class operator is a scaled
orthogonal projector, the selection criteria of three of the seven
variants become provably arbitrary or provably void, and the truncated
pseudo-inverse used to implement them removes exactly the subspace that
contains all between-class information (Section~\ref{sec:method}).
Replacing the truncation with a ridge term removes the degeneracy and
places the seven variants on the boundary of a two-parameter regularized
family \citep{qu2026modular}. The rest of the answer is empirical. Our
contributions are:

\begin{enumerate}
    \item \textbf{An exact, closed-form diagnosis} of how balanced
    sampling makes the truncated implementation degenerate. All signal
    eigenvalues of two variants are equal, and the operator of a third
    variant is exactly zero. We verify both results to floating-point
    precision on frozen sentence embeddings and, separately, on
    residual-stream activations of a decoder-only generative model. We
    also show a practical consequence: the accuracies of the truncated
    variants change with the number of BLAS threads, by up to 12
    percentage points, and by 57 for the variant whose operator is zero
    (Sections~\ref{sec:method:degeneracy}
    and~\ref{sec:results:reproducibility}).

    \item \textbf{A four-dataset, three-embedding-model evaluation} of the
    regularized formulation, with the ridge selected within the support
    set. It is compared with the seven original variants, a properly
    cross-validated linear probe, nearest centroid, shrinkage LDA, and
    three trained or adapted baselines (SetFit, LoRA, in-context
    learning). The regularized formulation outperforms the best original
    variant in 45 of 48 settings, but in general it does not outperform
    the probe or nearest centroid on the two bidirectional encoders
    (Section~\ref{sec:results:headline}).

    \item \textbf{Identification of the component that helps.} The
    regularized formulation gains most on E5-Mistral, the only
    decoder-based embedding model, and the gain comes from regularized
    within-class whitening rather than from KLPCDA. Shrinkage LDA, which
    has this component and nothing else, outperforms the tuned probe and
    nearest centroid in all 16 E5-Mistral settings. Pre-specified tests
    show that the effect depends on the embedding model rather than on
    its dimension, and that it becomes smaller as $k$ increases
    (Section~\ref{sec:results:whitening}).

    \item \textbf{Five guidelines} for practitioners, each limited to the
    conditions tested (Section~\ref{sec:guidelines}), and a revised
    interpretation of three geometric separability metrics. The raw
    metrics do not predict accuracy across encoders. A whitened metric
    that ranked E5-Mistral first is partly supported by the results of
    the classifiers that whiten (Section~\ref{sec:discussion}).
\end{enumerate}

\paragraph{Relation to the first version.} The first version of this
paper evaluated only the truncated implementation, repaired two variants
with a tie-break, and concluded that a tuned probe outperforms every
KLPCDA variant on three of four datasets. The diagnosis is unchanged. The
conclusion holds for the original variants, but in general not for the
regularized formulation. The first version's advice on which variant to
prefer for a given number of classes reflected the cost of truncation,
and we withdraw it (Section~\ref{sec:results:classcount}). The results
for the original variants are those of the first version; we recomputed
and checked them before running any new method (Section~\ref{sec:setup}).

We release the estimator, the evaluation code, and the per-seed results
for every table and figure.

%% file: sections/related_work.tex
\paragraph{Small-sample discriminant analysis.} Fisher's linear
discriminant analysis \citep{fisher1936use} and its many small-sample
extensions address the case $n < d$, where the within-class scatter
matrix $S_w$ is singular or ill-conditioned and standard LDA is undefined
(see \citet{qu2024review} for a recent survey of this problem and the
class-level and robustness challenges that accompany it). One line of
work kernelizes the discriminant problem, projecting onto directions
found by an orthogonal transformation in a reproducing kernel Hilbert
space rather than operating in the original feature space directly
\citep{qu2023rkhs}. Another regularizes the within-class covariance
instead: regularized discriminant analysis \citep{friedman1989regularized}
shrinks it toward a scaled identity, the Ledoit--Wolf estimator
\citep{ledoit2004well} chooses the amount of shrinkage analytically, and
the kernel Fisher discriminant \citep{mika1999fisher} adds a ridge to the
within-class term in the dual. KLPCDA \citep{qu2026klpcda} extends this kernelized
approach into a seven-variant framework that fuses total variance,
between-class scatter, and within-class scatter in different
combinations, allowing the $S_w$ term to be excluded entirely when it
is unreliable. A companion study \citep{qu2026modular} derives the structure of such fused
objectives when the number of samples is smaller than the dimension, places
the seven variants on the boundary of a two-parameter regularized family,
and finds on five classical small-sample domains (hyperspectral imaging,
bearing-fault vibration signals, gene expression, face and digit recognition)
that a fixed regularized member is preferable to choosing among the
variants. This paper studies the balanced $k$-shot setting on text
embeddings from pretrained language models.

\paragraph{Frozen sentence embeddings.} Sentence-transformer models
\citep{reimers2019sentencebert} produce fixed-length embeddings optimized
so that semantic similarity corresponds to embedding-space proximity,
spanning a wide range of sizes and architectures: small distilled
bidirectional encoders such as MiniLM \citep{wang2020minilm}
($d=384$), larger bidirectional encoders such as BGE
\citep{xiao2023cpack} ($d=1024$), and, more recently, billion-parameter
decoder-only language models repurposed as embedders via contrastive
fine-tuning, such as E5-Mistral \citep{wang2024e5mistral} ($d=4096$). The
MTEB benchmark \citep{muennighoff2023mteb} tracks these models' performance
on retrieval, clustering, and classification tasks using the full
available training data; we instead study the $k$-shot regime, where only
a handful of labeled examples per class are available, a setting MTEB's
own classification protocol does not test.

\paragraph{Few-shot text classification baselines.} We compare against
three baselines that adapt an encoder or model to the $k$-shot task
directly. SetFit \citep{tunstall2022setfit} fine-tunes a sentence
transformer via contrastive learning on pairs drawn from the support set,
then fits a lightweight classification head, a metric-learning approach
that does not require prompts. LoRA \citep{hu2022lora} instead
parameter-efficiently fine-tunes a standard classification head end-to-end
via low-rank adapters and ordinary cross-entropy loss. In-context learning
\citep{brown2020gpt3} places labeled examples directly in an instruction-following
language model's prompt and classifies by generation, with no
gradient update at all. These three baselines span the space from
metric-learning adaptation, through parameter-efficient fine-tuning, to
no adaptation at all, and the relative performance of these methods
depends systematically on $k$ (Section~\ref{sec:guidelines}).

\paragraph{Embedding anisotropy.} Prior work has documented that
contextualized and sentence embedding spaces are not isotropic:
representations cluster in a narrow cone rather than spreading uniformly
\citep{ethayarajh2019contextual}. We test whether anisotropy, or a
related linearly-separability-aware measure, explains why one embedding
model underperforms another in our $k$-shot setting; unlike prior work,
which studies anisotropy as a property of an embedding space in isolation,
we test it as a \emph{predictor} of downstream few-shot accuracy across
architecturally different encoders. Raw anisotropy does not predict it,
but the classifiers that whiten by a regularized within-class covariance
gain most on the encoder that distance-based rules handle worst
(Section~\ref{sec:discussion}).

%% file: sections/method.tex
\subsection{The Seven KLPCDA Variants}
\label{sec:method:variants}

KLPCDA is a supervised dimensionality reduction: it learns a projection
from labeled data, and a separate classifier then operates on the
projected features. Given $n$ training examples with kernel matrix $K$
(double-centered to $\tilde K$) and class labels over $L$ classes, KLPCDA
defines a between-class scatter operator $B$ and a within-class scatter
operator $W$ (both $n \times n$, acting on the dual/kernel-coefficient
space) and combines them with the (centered) total-variance operator
$C = \tilde K/n$ into seven target functions, summarized in
Table~\ref{tab:variants}. Each variant solves for the top (or, for No.7,
bottom) eigenvectors of an operator $E$ built from $B$, $W$, and $C$,
projects the data onto those eigenvectors, and classifies with
1-nearest-neighbor, following the evaluation convention of
\citet{qu2026klpcda}.

\input{tables/variants_table.tex}

\subsection{An Exact Eigenvalue Degeneracy Under Balanced $k$-Shot Sampling}
\label{sec:method:degeneracy}

No.1, No.3, and No.5 all require $\mathrm{pinv}(W\tilde K)$, and $W\tilde K$
is always singular. An early version of our estimator added a small ridge
before the pseudo-inverse. We replaced it with a truncated pseudo-inverse,
which removes the null space of $W$, because the ridge made No.1, No.3
and No.5 return identical accuracies. The analysis below, together with
its regularized counterpart in Section~\ref{sec:method:regularized},
shows that this interpretation was wrong. The identical accuracies were
the expected behaviour of a vanishing ridge, and the null space that
truncation removes contains all of the between-class information. The
first version of this paper evaluated the truncated implementation, and
this subsection analyzes it. After the change, No.1 and No.5 were still
among the least reliable variants, and their accuracy was unstable from
seed to seed. We derive the exact cause in closed form.

\paragraph{$W$ is an exact scaled projector.} For a class-$c$ block of
size $n_c$, $W$'s block is $p_c I_{n_c} - \tfrac{1}{n}\mathbf{1}_{n_c}\mathbf{1}_{n_c}^\top$
with $p_c = n_c/n$. Under \emph{balanced} $k$-shot sampling ($n_c = k$ for
every class, $n = kL$), $p_c = 1/L$ for every class, and each block
simplifies to $\tfrac{1}{L}\big(I_k - \tfrac1k \mathbf{1}_k\mathbf{1}_k^\top\big)$,
$1/L$ times the standard per-class centering projector. Stacking
blocks,
\begin{equation}
    W = \tfrac{1}{L}\,\Pi,
    \label{eq:w-projector}
\end{equation}
where $\Pi$ is the block-diagonal orthogonal projector that centers each
class separately. A projector has exactly two eigenvalues, so $W$ has
exactly two: $0$ with multiplicity $L$ (the per-class-mean directions,
$\Pi$'s null space) and $1/L$ with multiplicity $n-L$ (everything
orthogonal to those directions). We verified this in closed form and
against real embeddings (Banking77, $k=5$, $L=77$) to float64 precision:
$W$'s eigenvalues were exactly $\{0 \text{ (mult. } 77), 1/77 \text{ (mult.
} 308)\}$.

\paragraph{$B$ lives entirely in $\Pi$'s null space.} By the same kind of
argument, $B$'s blocks are built entirely from class-indicator outer
products, which span exactly the $L$-dimensional subspace $\Pi$
projects \emph{away}. Restricting $B$ to that subspace and diagonalizing
gives eigenvalues $0$ (multiplicity 1, the global-mean direction) and
$1/(kL)$ (multiplicity $L-1$, the between-class-contrast directions),
and $B$ is exactly zero on $\Pi$'s $(n-L)$-dimensional range. We verified
$\Pi B \Pi = 0$ directly (max absolute value $2.8\times10^{-36}$ on real
data).

\paragraph{Consequences for No.1, No.3, No.5.} Chaining these two facts
through $\mathrm{pinv}(W\tilde K)$:

\begin{itemize}
    \item \textbf{No.5} ($E = \mathrm{pinv}(W\tilde K)\,\tilde K$): every
    one of the $n-L$ ``signal'' eigenvalues of $E$ is \emph{exactly} $L$ (we
    measured $77.000000000\pm1.2\times10^{-10}$ on real data). No.5's own
    selection criterion is therefore \emph{provably indifferent} across its
    entire signal subspace. \texttt{numpy.linalg.eig} returns an
    arbitrary orthonormal basis of a tied eigenspace, and this basis
    determines No.5's final $n_{\text{components}}$ directions.
    \item \textbf{No.3} ($E = \mathrm{pinv}(W\tilde K)\,(B\tilde K)$):
    since $B\tilde K$'s output lies entirely in the subspace
    $\mathrm{pinv}(W\tilde K)$ cannot recover any signal from, exactly
    $\Pi$'s null space where $W$ is zero, $E$ is provably the zero
    matrix. We measured eigenvalues at $10^{-12}$ to $10^{-14}$, floating-point
    noise around an exact zero, not a small real signal. No.3's objective
    is \emph{void} under balanced sampling, independent of embedding
    quality or dataset.
    \item \textbf{No.1} ($E = \mathrm{pinv}(W\tilde K)(\alpha C + \beta B\tilde K)$,
    default $\alpha=\beta=1$): since the $B$-term contributes exactly
    nothing, No.1 collapses to $\tfrac{1}{n}\cdot L = 1/k$ exactly on its
    signal subspace, fully explaining a previously unexplained
    observation that No.1's top eigenvalues were ``suspiciously equal to
    almost exactly $1/k$.''
\end{itemize}

No.7 ($E = W\tilde K$, selecting the \emph{smallest} eigenvalues) selects
$W$'s null space directly, also exactly degenerate (all $L$ eigenvalues
at $0$), but this subspace is spanned by the class-indicator vectors
themselves, so it is structurally close to a kernelized nearest-centroid
rule, which plausibly explains why No.7 is degenerate but less unstable
than No.1/No.3/No.5.

The structure above is the balanced special case of a general one: for
arbitrary class sizes, the range of $B$ lies in the null space of $W$, the
finite eigenvalues of No.5 under a truncated pseudo-inverse are $1/n_c$ with
multiplicity $n_c-1$, and the seven variants lie on the boundary of a regularized
family with two parameters \citep{qu2026modular}.

\subsection{A Tie-Break for the Truncated Implementation}
\label{sec:method:fix}

In the truncated implementation, the criterion of No.5 (and of No.1)
cannot distinguish between directions within its degenerate subspace.
The first version of this paper therefore used each variant's own
secondary term to break the tie (for No.5, its $C$ term; for No.1, its
fused numerator $\alpha C + \beta B$). Within any run of near-equal
eigenvalues of $E$ (relative tolerance $10^{-3}$), we re-orthonormalize
the tied eigenvectors and diagonalize the tie-break matrix restricted to
their span. Ranking the tied directions by \emph{ascending} variance
improved accuracy (Banking77, $k{=}5$: $0.81$, compared with $0.54$ for
the arbitrary basis), whereas the literal reading of No.5's ``maximize''
(descending order) reduced it ($0.15$). Section~\ref{sec:discussion}
returns to this open question. The tie-break is only a partial fix. It
chooses a basis within a block whose directions carry no between-class
information on the training set, and it does not fix No.3 or No.7,
which have no comparable secondary term. The regularized formulation
below removes the degeneracy rather than working around it.

\subsection{The Regularized Formulation}
\label{sec:method:regularized}

Replacing the denominator $v^\top S_w v$ with $v^\top S_w v + \mu\|v\|^2$
(a ridge, $\mu>0$) turns each ratio-type variant into a symmetric,
positive-definite generalized eigenproblem in the coordinates of the
range of the centered kernel. No pseudo-inverse is then needed, and the
formulation itself no longer forces any eigenvalues to be equal.
\citet{qu2026modular} shows that, with the fusion ratio
$t=\beta/\alpha$, the variants become the corners and edges of a
two-parameter family (Table~\ref{tab:variants}). As $\mu\to0^{+}$, the
leading $L-1$ directions converge to the null space of $W$ with a
uniquely determined basis, the null-space discriminant directions. This
assumes $\mathrm{rank}\,\tilde K = n-1$; when the rank $r$ is lower, as
in several of our settings, the null space shrinks to $\max(0, r-n+L)$
dimensions. As $\mu\to\infty$, the family reduces to No.2 (and to No.4
or No.6 at the ends of $t$). The truncated implementation above is the
singular limit $\mu=0$ with the null-space block removed. The same study
finds that the fusion ratio $t$ does not improve accuracy, whereas $\mu$
does, so we use the two edges of the family:
\textbf{RKFD}, the regularized kernel Fisher discriminant
\citep{mika1999fisher} (numerator $S_b$, i.e.\ No.3 with a ridge), and
\textbf{No.5$_\mu$} (numerator $C$). With a linear kernel, RKFD is
ridge-regularized LDA expressed in the dual
\citep{friedman1989regularized}.

The ridge is selected using the support set only. Two scales are
computed from the kernel of each fit: $\mu_s$, below which the
null-space directions dominate, and $\mu_l$, above which the solution
approaches No.2. The grid has 13 values spaced logarithmically from
$\mu_s/100$ to $100\,\mu_l$, so the selected quantity is a scale-free
grid index. The index is selected by stratified cross-validation within
the support set, using the same splitter that the logistic-regression
probe uses to select its regularization (Section~\ref{sec:setup}), with
1-nearest-neighbor accuracy on the held-out folds as the score. Ties are
broken in favour of the index closest to the middle of the grid, and
then the larger index. As for the original variants, 1-nearest-neighbor
classification is applied after the projection. The subspace dimension
follows the original protocol: $\min(30, L-1)$ for RKFD (the rule the
first version used for No.3 and No.6) and $\min(30, n-1)$ for
No.5$_\mu$. As a pre-specified sensitivity analysis, we also report all
projection methods, original and regularized, without the cap of 30.

\subsection{Shrinkage LDA}
\label{sec:method:lda}

To separate what is specific to KLPCDA from what any regularized
within-class whitening provides, we add a standard baseline that has this
component and nothing else: linear discriminant analysis with a shrunk
within-class covariance. The covariance is estimated from the pooled
within-class residuals of the support set (each example minus its class
mean) with Ledoit--Wolf shrinkage \citep{ledoit2004well}, and it is
shared by all classes. Classification uses the linear discriminant rule
with equal priors, since the support set is balanced. We use this pooled
form rather than shrinking the covariance of each class separately, as
some library implementations do, because the per-class form degenerates
with two examples per class.

%% file: tables/variants_table.tex
\begin{table}[t]
\centering
\caption{The seven KLPCDA variants. $C$ = total variance, $S_b$ =
between-class scatter, $S_w$ = within-class scatter. ``$S_w$ involved''
marks the four variants (No.1, No.3, No.5, No.7) whose target function
involves $W$ and which are therefore subject to the degeneracy of
Section~\ref{sec:method:degeneracy}: No.1/No.3/No.5 through
$\mathrm{pinv}(W\tilde K)$, No.7 through $W\tilde K$ directly. The last
column gives each variant's position in the ridge-regularized family of
\citet{qu2026modular} (Section~\ref{sec:method:regularized}), with
$\mu$ the ridge and $t=\beta/\alpha$ the fusion ratio; ``$\mu=0$'' means
the truncated pseudo-inverse, which removes the null space of $W$.}
\label{tab:variants}
\begin{tabular}{@{}llccl@{}}
\toprule
Method & Fused terms & Target function & $S_w$ involved & Position in $(\mu, t)$ \\
\midrule
No.1 & $C, S_b, S_w$ & $(\alpha C + \beta S_b)v \,/\, S_w v$ & yes & $\mu=0$, ratio $t$ \\
No.2 & $C, S_b$       & $\alpha Cv + \beta S_b v$             & no  & $\mu\to\infty$, ratio $t$ \\
No.3 & $S_b, S_w$     & $S_b v \,/\, S_w v$                   & yes & $\mu=0$, $t\to\infty$ (RKFD for $\mu>0$) \\
No.4 & $C$             & $Cv$ (centered KPCA)                 & no  & $\mu\to\infty$, $t=0$ \\
No.5 & $C, S_w$        & $Cv \,/\, S_w v$                     & yes & $\mu=0$, $t=0$ (No.5$_\mu$ for $\mu>0$) \\
No.6 & $S_b$           & $S_b v$                               & no  & $\mu\to\infty$, $t\to\infty$ \\
No.7 & $S_w$           & $\mathrm{argmin}\; S_w v$              & yes & outside (order independent of $\mu$) \\
\bottomrule
\end{tabular}
\end{table}

%% file: sections/setup.tex
\subsection{Datasets}

We evaluate on four public text classification benchmarks, chosen to
vary in class count by nearly two orders of magnitude: \textbf{Banking77}
(77 fine-grained banking intents), \textbf{CLINC150} (150 intents; its
out-of-scope class is excluded), \textbf{TREC} (6 coarse question-type
classes), and \textbf{AG News} (4 topic classes). This range matters:
several findings observed on the two many-class datasets did not
replicate on the two few-class datasets (Section~\ref{sec:results}).

\subsection{Embedding Models}

We extract frozen embeddings at three dimensions: \textbf{MiniLM}
\citep{wang2020minilm} ($d=384$, mean-pooled, final layer), \textbf{BGE-large}
\citep{xiao2023cpack} ($d=1024$), and \textbf{E5-Mistral-7B-Instruct}
\citep{wang2024e5mistral} ($d=4096$, a decoder-only instruction-tuned
7B-parameter LLM embedder run in float16). MiniLM and BGE-large are
bidirectional encoders; E5-Mistral is the only decoder-based embedding
model in the study. No embedding model is fine-tuned; downstream methods
use only the labels in the $k$-shot support set.

\subsection{$k$-Shot Protocol}

For each dataset and $k \in \{2,3,5,10\}$, we draw 10 independent
stratified random support sets, each containing exactly $k$ examples per
class sampled without replacement, and evaluate every method on the full
official test split. Here, ``few-shot'' means a supervised method trained
on these $k$ examples per class on top of a frozen embedding; in-context
learning, which the term often refers to for language models, is one of
our baselines. We report means over the 10 seeds and use paired Wilcoxon
signed-rank tests, matched by seed, to compare methods within a setting.
One analysis goes beyond this range: the estimation-efficiency analysis
in Section~\ref{sec:discussion} extends $k$ to $20$--$50$, subject to the
size of each dataset's smallest class.

The results for the original variants, the probe, and nearest centroid
are those reported in the first version of this paper. Before computing
any new method, we recomputed these methods on two seeds at every $k$ in
every setting and required agreement with the stored values to within
$10^{-9}$. All runs use the default number of BLAS threads (16 on our
machine); Section~\ref{sec:results:reproducibility} explains why this
matters. For the comparisons involving the regularized formulation and
shrinkage LDA, the decision rules and analysis scripts were specified and
recorded before the runs.

\subsection{Methods Compared}

\textbf{Frozen-embedding methods}: the seven original KLPCDA variants
(linear kernel, truncated pseudo-inverse, tie-break for No.1 and No.5,
1-nearest-neighbor on the projected features); the regularized RKFD and
No.5$_\mu$ (Section~\ref{sec:method:regularized}); shrinkage LDA
(Section~\ref{sec:method:lda}); a logistic regression probe with $C$
selected by cross-validation within the $k$-shot support set only; and a
nearest-centroid classifier. The test set is not used for model
selection. An earlier evaluation using the default $C=1$ underestimated
the probe's accuracy by 3--4 percentage points at every setting, which
overstated the relative performance of every KLPCDA variant.

\textbf{Trained/adapted baselines}: SetFit \citep{tunstall2022setfit}
(contrastive fine-tuning of the MiniLM encoder followed by fitting a
lightweight head), LoRA \citep{hu2022lora} (rank-16 low-rank adapters and
a new classification head trained with cross-entropy, with
hyperparameters selected through a small sweep after the default
configuration underperformed by more than 20 accuracy points), and
in-context learning with \textbf{Qwen2.5-3B-Instruct}, an open
instruction-tuned model run locally without an external API. ICL places
the $k$-shot support set in the prompt as demonstrations and classifies
by constrained generation. We evaluate ICL only on TREC and AG News,
because including demonstrations for all classes in a single prompt is
impractical for datasets with 77--150 classes; for example, CLINC150 at
$k{=}10$ would require more than 1{,}500 demonstrations per prompt. We
return to this limitation in Section~\ref{sec:discussion}.

%% file: sections/results.tex
\subsection{Main Result: The Regularized Formulation Outperforms the Original Variants, but Not the Strong Baselines}
\label{sec:results:headline}

Table~\ref{tab:summary} gives counts over the 48 settings (four datasets,
three embedding models, four values of $k$). RKFD outperforms the best
of the seven original variants in 45 of them (significantly, by paired
Wilcoxon tests over seeds, in 30), and in 46 without the subspace cap.
The best original variant is No.6 in nearly every setting. The
regularized formulation is therefore a clear improvement over the
original variants.

The comparison with the two strong frozen-embedding baselines is mixed.
RKFD matches or exceeds the tuned probe in 21 of 48 settings (26 without
the cap); the probe is significantly better in 21 settings and RKFD in 11.
Under the original protocol, the probe outperforms RKFD in every setting
with $k\ge3$ only on CLINC150; without the cap, this holds on no dataset,
and RKFD matches or exceeds the probe in all 12 TREC settings under both
rules. Nearest centroid, a training-free baseline, outperforms RKFD in 31
of 48 settings. Shrinkage LDA, which shares RKFD's regularized
within-class whitening and nothing else, matches or exceeds RKFD in 30
settings (by 1.0 point on average), the probe in 26, and nearest centroid
in 20. In the single slice shown in Table~\ref{tab:headline} ($k=5$,
MiniLM), the probe is best on three datasets and shrinkage LDA on TREC.

The first version of this paper concluded that a properly tuned probe
outperforms every KLPCDA variant on Banking77, CLINC150 and AG News at
every $k\ge3$. This conclusion was a consequence of the truncated
implementation. It holds for the original variants, but for the
regularized formulation it holds only on CLINC150 and only with the
subspace cap. Section~\ref{sec:results:whitening} examines where the
regularized methods gain most.

\begin{figure}[t]
\centering
\includegraphics[width=\linewidth]{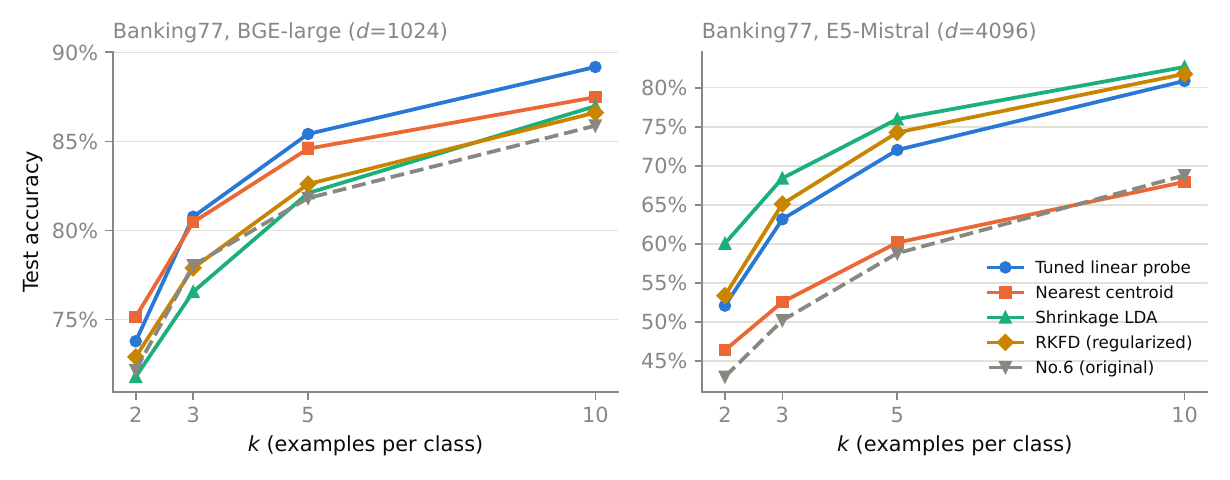}
\caption{Banking77 with a bidirectional encoder (left, BGE-large) and with
the decoder-based embedding model (right, E5-Mistral), mean test accuracy
over 10 seeds. With BGE-large, the tuned probe and nearest centroid
perform best, and the regularized methods are close to the original No.6.
With E5-Mistral, nearest centroid and No.6 fall far behind, and shrinkage
LDA and RKFD outperform the probe.}
\label{fig:landscape}
\end{figure}

\subsection{Numerical Reproducibility}
\label{sec:results:reproducibility}

Running exactly the same computation with 16, 8 or 1 BLAS threads changes
the reported accuracy of the truncated variants (Table~\ref{tab:threads}):
by up to $0.070$ for No.1, $0.118$ for No.5 and $0.082$ for No.7, and by
up to $0.571$ for No.3, whose operator is exactly zero
(Section~\ref{sec:method:degeneracy}). Changing the number of threads
changes the order of floating-point reductions. For operators that are
singular by definition, or whose leading eigenvalues are exactly equal,
this is enough to move the truncation boundary or change the returned
basis. The stored results of the first version are reproduced exactly
only with the default 16 threads of our machine. RKFD, No.5$_\mu$, the
probe, nearest centroid, and the variants without $S_w$ (No.2, No.4,
No.6) give identical accuracies under all three settings in all twelve
dataset--embedding combinations, since a symmetric positive-definite
problem does not have this sensitivity. The accuracies reported for the
truncated variants should therefore be read as one sample from a range
that depends on the implementation.

\subsection{Where Regularized Whitening Helps: E5-Mistral}
\label{sec:results:whitening}

The regularized methods differ most from the other frozen-embedding
methods on E5-Mistral (Figure~\ref{fig:landscape}, right). In its 16
settings, shrinkage LDA outperforms the tuned probe in all 16 (by $3.1$
points on average) and nearest centroid in all 16 (by $7.1$); RKFD
outperforms the probe in 11 and nearest centroid in 14. On the two
bidirectional encoders, the opposite holds: RKFD outperforms the probe in
10 of 32 settings, mostly on TREC, and nearest centroid in 3, and
shrinkage LDA is below nearest centroid in all 16 BGE-large settings.

We noticed this pattern only after the comparison above, so we tested it
with criteria fixed in advance, using a method, data and a condition that
were not used to find it. The measured quantity is how much more a method
gains over nearest centroid on E5-Mistral than on BGE-large, for the same
dataset and $k$. (i) With shrinkage LDA in place of RKFD, the gain is
larger on E5-Mistral in all 16 dataset--$k$ pairs, by $8.6$ points on
average. (ii) On E5-Mistral embeddings projected by PCA to $d=1024$, the
dimension of BGE-large, RKFD's gain remains larger in 7 of 8 pairs (by
$6.1$ points), and shrinkage LDA's in 8 of 8. (iii) On Banking77 with
larger support sets, RKFD's additional gain on E5-Mistral decreases from
$0.147$ at $k=10$ to $0.125$ at $k=20$ and $0.114$ at $k=30$. The effect
is therefore a property of regularized within-class whitening rather
than of KLPCDA; it depends on the embedding geometry of E5-Mistral rather
than on its dimension; and it becomes smaller as more labeled examples
are available. A plausible interpretation is that the raw Euclidean
geometry of E5-Mistral is poorly suited to distance-based rules such as
nearest centroid, and that whitening with a well-conditioned
within-class covariance corrects much of this. We did not measure the
geometry directly, and E5-Mistral is the only decoder-based embedding
model in the study.

\subsection{Embedding Dimension}
\label{sec:results:dimension}

Table~\ref{tab:dreversal} shows Banking77 at $k=5$ across embedding
models. For the original variants and the probe, accuracy drops at
$d{=}4096$ and the probe's lead over the original variants increases
(first version: paired Wilcoxon $p{=}0.002$, 10 of 10 seeds, on both
many-class datasets at every $k\le10$). The regularized methods lose much
less and outperform the probe on Banking77 (RKFD $0.743$, shrinkage LDA
$0.760$, probe $0.720$); on CLINC150 the probe remains ahead of RKFD with
the subspace cap, but not without it. Projecting the same E5-Mistral
embeddings to $d{=}1024$ leaves the probe unchanged ($0.719$) and the
regularized methods ahead ($0.730$ and $0.754$), so the behaviour depends
on the embedding model, not on $d$ itself. Every off-the-shelf embedding
model above $d{=}1024$ that we could find is a decoder-only,
instruction-tuned LLM embedder, so $d$ cannot currently be varied
independently with a bidirectional encoder. The first version also
reported that increasing $d$ from 384 to 1024 narrows the gap between
the probe and the best variant on the two many-class datasets; that
observation concerned the original variants, and we did not re-examine
it for the regularized ones.

\subsection{The Cost of Truncation and Class Count}
\label{sec:results:classcount}

\begin{figure}[t]
\centering
\includegraphics[width=0.6\linewidth]{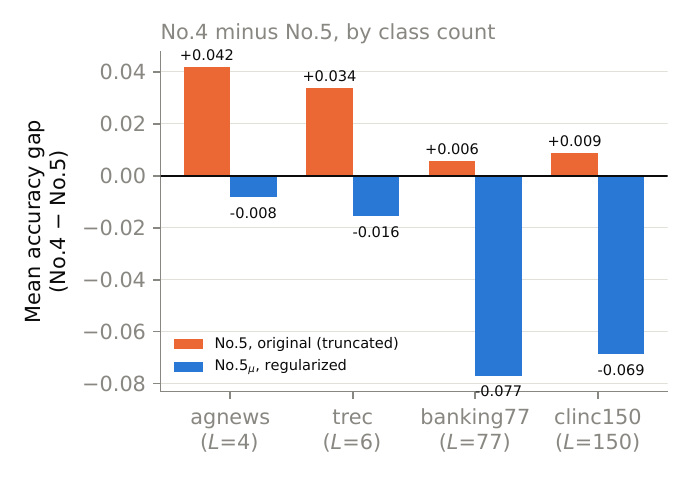}
\caption{No.4 minus No.5, averaged over $k\in\{2,3,5,10\}$ and three
embedding models, for the original truncated No.5 (orange) and the
regularized No.5$_\mu$ (blue). Truncation costs most on the few-class
datasets; removing it reverses the sign everywhere, most strongly on the
many-class datasets.}
\label{fig:classcount}
\end{figure}

In the truncated implementation, No.5 is restricted to the subspace
orthogonal to the class-mean directions. The first version measured the
cost of this restriction as the gap between No.4 (the same variance
criterion without the restriction) and No.5. Pooled over embedding models
and $k$, the gap is $0.0072$ on the two many-class datasets and $0.0379$
on the two few-class datasets (Mann--Whitney test over the 24 + 24
settings, $p = 1.6\times10^{-6}$; the first version printed this as
$0.000001$). The regularized No.5$_\mu$ is not restricted to that
subspace, and the sign of the gap reverses
(Figure~\ref{fig:classcount}): No.5$_\mu$ is above No.4 in 46 of 48
settings (all 24 many-class settings and 22 of 24 few-class settings), by
$0.073$ on the many-class datasets and $0.012$ on the few-class ones
($p = 6.6\times10^{-6}$). Class count still affects the gap, but in the
opposite direction to the first version's interpretation. This is
consistent with the fact that the null-space block removed by truncation
has $L-1$ dimensions, and therefore carries more of the class information
when $L$ is large.

\subsection{Layer, Pooling, and the Margin Metric Within One Model}
\label{sec:results:layers}

\begin{figure}[t]
\centering
\includegraphics[width=0.55\linewidth]{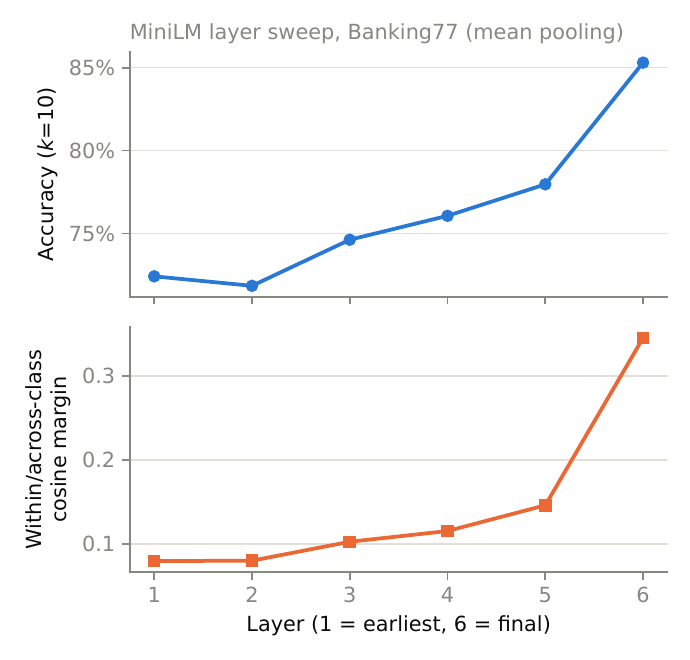}
\caption{Layer sweep, MiniLM on Banking77. The within/across-class margin
(bottom) follows accuracy (top) almost exactly across the layers of the
same model, a relationship that does not hold across different encoder
architectures (Section~\ref{sec:discussion}).}
\label{fig:layers}
\end{figure}

A full layer sweep of MiniLM on Banking77 (Figure~\ref{fig:layers}, top)
shows that accuracy increases monotonically with depth, with a large jump
at the final layer. This is expected for a model trained end-to-end with
a final-layer, mean-pooling contrastive objective, and the sweep confirms
it with measurements. Mean pooling outperforms CLS and max pooling at
every $k$, consistent with MiniLM's own default configuration.

We had previously found that a simple within-class-versus-across-class
cosine margin does not predict accuracy \emph{across} architecturally
different encoders (MiniLM vs.\ BGE-large vs.\ E5-Mistral): MiniLM has
the largest margin of the three but the lowest accuracy. When computed
\emph{within} the layers of one model (Figure~\ref{fig:layers}, bottom),
the same margin metric follows accuracy almost exactly, with the same
monotonic ordering. The metric is therefore informative for comparing
configurations of the same model, but it
cannot be compared across different architectures without further
adjustment.

\subsection{Trained Baselines: SetFit, LoRA, and In-Context Learning}
\label{sec:results:baselines}

On MiniLM, whose encoder SetFit fine-tunes, SetFit outperforms the best
original variant in 15 of 16 settings (the exception is Banking77 at
$k{=}10$, $0.801$ compared with $0.803$; the first version stated all 16),
RKFD in 13, and shrinkage LDA in 11. It also outperforms the tuned linear
probe at $k{=}2$ on Banking77 and CLINC150, but falls behind as $k$
increases, which is expected for a metric-learning method with a strong
prior but limited capacity to use additional labels. LoRA performs worse
than SetFit at every $k$ on Banking77 and CLINC150. This is consistent
with the literature: training a new $L$-way softmax head from scratch is
a harder optimization problem than SetFit's contrastive approach,
especially with 77--150 classes. However, this pattern does not
generalize to TREC and AG News, where the relative performance of SetFit
and LoRA crosses over with $k$ in opposite directions on the two
datasets. We checked directly whether class count explains this
crossover, since it explains much else in this paper; it does not. AG
News has four classes, fewer than TREC's six, yet shows the opposite
crossover. We report this as an unresolved, dataset-specific finding
rather than propose a second class-count explanation that the data do
not support.

\begin{figure}[t]
\centering
\includegraphics[width=0.85\linewidth]{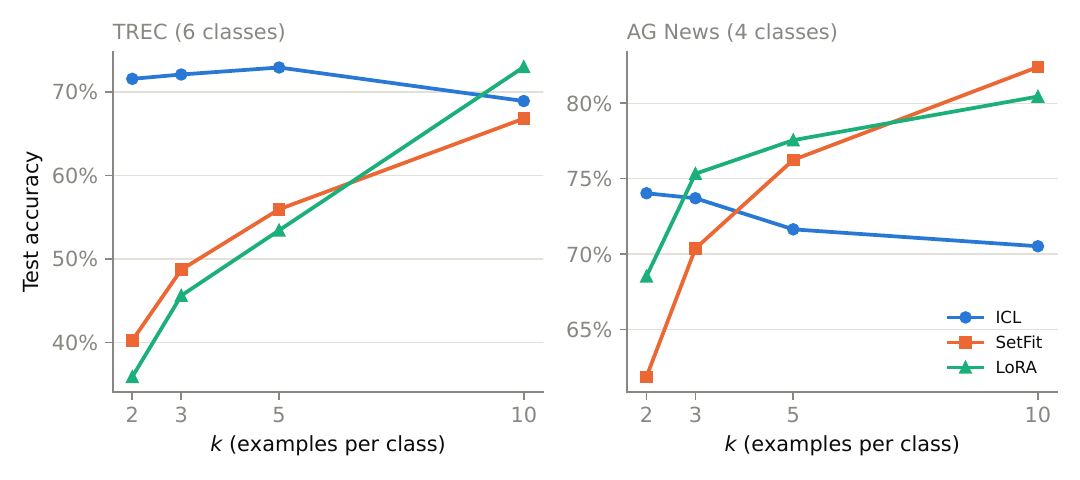}
\caption{The three trained/adapted baselines on TREC and AG News, the
only datasets on which all three were evaluated. ICL's advantage over
both fine-tuned methods is concentrated at $k{\le}3$.}
\label{fig:baselines}
\end{figure}

In-context learning with Qwen2.5-3B-Instruct, evaluated only on TREC and
AG News (Section~\ref{sec:setup}), shows a clearer pattern
(Figure~\ref{fig:baselines}). It outperforms both fine-tuned baselines
by a wide margin at $k{=}2,3$ on both datasets. By $k{=}10$, at least one
of SetFit or LoRA has caught up with or overtaken it on each dataset.
This is the clearest example in our experiments of a pattern that
appears throughout this study: methods that require no or little
fine-tuning have an advantage at the lowest shot counts, and this
advantage decreases or reverses once enough labeled data are available
for a fine-tuned method to fit well.

Table~\ref{tab:cost} reports wall-clock fit-plus-predict time on TREC
($k{=}5$), the only dataset on which every method family was run. The
frozen-embedding methods finish in well under a tenth of a second on CPU;
RKFD's time ($0.06$\,s) includes the cross-validation over $\mu$, and
shrinkage LDA takes $0.01$\,s. SetFit and LoRA require several seconds of
GPU fine-tuning per fold. In-context learning is both the slowest and the
most variable method, at $48.9\pm40.0$ seconds, because it has a
per-example generation cost at test time rather than a one-time training
cost. The regularized methods are therefore two to three orders of
magnitude cheaper than the fine-tuned baselines.

\input{tables/headline_table.tex}
\input{tables/summary_table.tex}
\input{tables/threads_table.tex}
\input{tables/d_reversal_table.tex}
\input{tables/estimation_efficiency_table.tex}
\input{tables/cost_table.tex}

%% file: tables/headline_table.tex
\begin{table}[t]
\centering
\caption{Test accuracy at $k=5$ on all four datasets (MiniLM, $d$=384), mean over 10 seeds. \textbf{Bold} marks the best method per dataset. RKFD is the ridge-regularized formulation with $\mu$ chosen by cross-validation inside the support set; No.5 is the original truncated-pseudo-inverse implementation with the tie-break of Section~\ref{sec:method:fix}.}
\label{tab:headline}
\begin{tabular}{@{}lcccc@{}}
\toprule
Method & Banking77 & CLINC150 & TREC & AG News \\
\midrule
Tuned linear probe & \textbf{0.797} & \textbf{0.871} & 0.472 & \textbf{0.716} \\
Nearest centroid & 0.768 & 0.865 & 0.526 & 0.715 \\
Shrinkage LDA & 0.771 & 0.861 & \textbf{0.529} & 0.698 \\
RKFD (regularized) & 0.763 & 0.826 & 0.526 & 0.700 \\
No.6 (original) & 0.744 & 0.820 & 0.506 & 0.688 \\
No.5 (original, tie-break) & 0.720 & 0.774 & 0.436 & 0.629 \\
\bottomrule
\end{tabular}
\end{table}

%% file: tables/summary_table.tex
\begin{table}[t]
\centering
\caption{Cell counts over all 48 settings (4 datasets $\times$ 3 embeddings $\times$ $k\in\{2,3,5,10\}$; each cell a mean over 10 seeds), 12 cells per dataset. ``Best orig.'' is the best of the seven original variants in that cell (an oracle choice, generous to them). RKFD uses the subspace cap of 30 of the original protocol; ``unc.'' removes the cap for every projection method.}
\label{tab:summary}
\begin{tabular}{@{}lcccccc@{}}
\toprule
& \multicolumn{4}{c}{RKFD} & \multicolumn{2}{c}{Shrinkage LDA} \\
\cmidrule(lr){2-5}\cmidrule(l){6-7}
Dataset & $>$ best orig. & $\geq$ probe & $\geq$ probe (unc.) & $\geq$ centroid & $\geq$ probe & $\geq$ centroid \\
\midrule
Banking77 & 11/12 & 5/12 & 6/12 & 5/12 & 5/12 & 6/12 \\
CLINC150 & 11/12 & 0/12 & 4/12 & 2/12 & 4/12 & 5/12 \\
TREC & 12/12 & 12/12 & 12/12 & 6/12 & 12/12 & 5/12 \\
AG News & 11/12 & 4/12 & 4/12 & 4/12 & 5/12 & 4/12 \\
\midrule
All & 45/48 & 21/48 & 26/48 & 17/48 & 26/48 & 20/48 \\
\bottomrule
\end{tabular}
\end{table}

%% file: tables/threads_table.tex
\begin{table}[t]
\centering
\caption{Numerical reproducibility. For each method, the largest change in test accuracy when the same computation is run with 16, 8 or 1 BLAS threads (seeds 0 and 1, every $k$), maximized over the twelve settings, and the number of settings in which that change exceeds 0.01.}
\label{tab:threads}
\begin{tabular}{@{}lcc@{}}
\toprule
Method & Max.\ change & Settings $>0.01$ \\
\midrule
No.1 (original, truncated pseudo-inverse) & 0.070 & 12/12 \\
No.3 (original, truncated pseudo-inverse) & 0.571 & 12/12 \\
No.5 (original, truncated pseudo-inverse) & 0.118 & 11/12 \\
No.7 (original, smallest eigenvalues) & 0.082 & 6/12 \\
RKFD & 0.000 & 0/12 \\
RKFD, uncapped & 0.000 & 0/12 \\
No.5$_\mu$ & 0.000 & 0/12 \\
No.5$_\mu$, uncapped & 0.000 & 0/12 \\
Tuned linear probe & 0.000 & 0/12 \\
Nearest centroid & 0.000 & 0/12 \\
No.2 (original) & 0.000 & 0/12 \\
No.4 (original) & 0.000 & 0/12 \\
No.6 (original) & 0.000 & 0/12 \\
\bottomrule
\end{tabular}
\end{table}

%% file: tables/d_reversal_table.tex
\begin{table}[t]
\centering
\caption{Banking77, $k=5$, across embeddings. The last column projects the same E5-Mistral embeddings to $d$=1024 by PCA. The regularized methods keep their advantage on E5-Mistral after the projection (Section~\ref{sec:results:whitening}).}
\label{tab:dreversal}
\begin{tabular}{@{}lcccc@{}}
\toprule
Method & $d$=384 & $d$=1024 & $d$=4096 (raw) & $d$=4096$\to$1024 (PCA) \\
\midrule
Tuned linear probe & 0.797 & 0.854 & 0.720 & 0.719 \\
Nearest centroid & 0.768 & 0.846 & 0.602 & 0.633 \\
No.6 (original) & 0.744 & 0.818 & 0.588 & 0.597 \\
RKFD (regularized) & 0.763 & 0.826 & 0.743 & 0.730 \\
Shrinkage LDA & 0.771 & 0.821 & 0.760 & 0.754 \\
\bottomrule
\end{tabular}
\end{table}

%% file: tables/estimation_efficiency_table.tex
\begin{table}[t]
\centering
\caption{Accuracy gap BGE-large minus E5-Mistral at matched $k$, per method. The probe's gap shrinks by more than 80\% from $k{=}2$ to the largest $k$ tested (Banking77 capped at $k{=}30$ by its smallest class). The regularized methods start from a smaller gap. `--' marks $k$ values not run.}
\label{tab:estefficiency}
\begin{tabular}{@{}lccccccc@{}}
\toprule
Dataset, method & $k{=}2$ & $k{=}3$ & $k{=}5$ & $k{=}10$ & $k{=}20$ & $k{=}30$ & $k{=}50$ \\
\midrule
Banking77, tuned probe & 0.217 & 0.176 & 0.134 & 0.083 & 0.050 & 0.040 & -- \\
Banking77, RKFD & 0.195 & 0.128 & 0.083 & 0.048 & 0.033 & 0.032 & -- \\
Banking77, shrinkage LDA & 0.117 & 0.081 & 0.061 & 0.043 & 0.030 & 0.025 & -- \\
CLINC150, tuned probe & 0.190 & 0.150 & 0.114 & 0.076 & 0.054 & 0.041 & 0.028 \\
\bottomrule
\end{tabular}
\end{table}

%% file: tables/cost_table.tex
\begin{table}[t]
\centering
\caption{Wall-clock time (fit + predict on the full test set, mean $\pm$ std over 10 seeds) at $k=5$ on TREC, the one dataset where every method family was run. The frozen-embedding methods ran CPU-only; SetFit, LoRA and ICL required GPU fine-tuning or per-example generation. RKFD's time includes the cross-validation over $\mu$; its standard deviation is dominated by the first call.}
\label{tab:cost}
\begin{tabular}{@{}lr@{}}
\toprule
Method & Seconds (fit + predict) \\
\midrule
Tuned linear probe & 0.03 $\pm$ 0.01 \\
Nearest centroid & 0.00 $\pm$ 0.00 \\
No.6 (original KLPCDA variant) & 0.00 $\pm$ 0.00 \\
RKFD (incl.\ selecting $\mu$) & 0.06 $\pm$ 0.10 \\
Shrinkage LDA & 0.01 $\pm$ 0.00 \\
SetFit (fine-tuned) & 7.19 $\pm$ 0.18 \\
LoRA (fine-tuned) & 3.29 $\pm$ 0.04 \\
In-context learning & 48.89 $\pm$ 39.95 \\
\bottomrule
\end{tabular}
\end{table}

%% file: sections/guidelines.tex
Each guideline below states a practical recommendation, the quantitative
evidence for it, and its mechanism where one is known. Each is limited to
the conditions tested; we do not generalize a partial result beyond the
available evidence.

\paragraph{Guideline 1 (do not choose among the original variants; use a
regularized formulation).} Under the standard balanced $k$-shot protocol,
the truncated implementations of No.1, No.3 and No.5 are exactly
degenerate (Section~\ref{sec:method:degeneracy}), and their accuracies
depend on the number of BLAS threads (Table~\ref{tab:threads}). A ridge
with $\mu$ selected within the support set removes both problems and
outperforms the best original variant in 45 of 48 settings
(Table~\ref{tab:summary}). The first version's advice to prefer No.2,
No.4 or No.6 over No.1 or No.5 when there are few classes reflected the
cost of truncation, and we withdraw it: the regularized No.5$_\mu$ is
above No.4 in 46 of 48 settings (Section~\ref{sec:results:classcount}).
This agrees with the cross-domain finding that choosing among the
variants does not pay off \citep{qu2026modular}. Caveat: the regularized
formulation improves on the original variants, but in general it does
not outperform the tuned probe or nearest centroid.

\paragraph{Guideline 2 (whiten where distance-based rules fail).} On
E5-Mistral, where nearest centroid is far behind the tuned probe, a
classifier with a regularized within-class covariance (shrinkage LDA or
RKFD) is the better choice: shrinkage LDA outperforms the probe in all
16 settings and nearest centroid in all 16
(Section~\ref{sec:results:whitening}). For another embedding model, a
large gap between nearest centroid and the probe in a pilot run is an
inexpensive indication that whitening may help, but we have not tested
this as a rule. On the bidirectional encoders tested (MiniLM, BGE-large),
nearest centroid or the probe is sufficient, and shrinkage LDA is below
nearest centroid in every BGE-large setting. Caveat: E5-Mistral is the
only decoder-based embedding model tested, and we did not directly
measure the geometric property that makes whitening effective.

\paragraph{Guideline 3 (the margin metric works within, but not across,
architectures).} When choosing between layers or pooling rules for an
encoder already in use, a within-class-to-across-class cosine margin
predicts relative accuracy reliably and at low cost, without a $k$-shot
pilot. When choosing between encoder architectures, it does not; the
actual comparison is necessary. Evidence: Figure~\ref{fig:layers}
(agreement within one model) and the MiniLM/BGE-large/E5-Mistral ranking
reversal in Section~\ref{sec:results} (failure across architectures).
Caveat: tested on one model (MiniLM) and one dataset (Banking77); not yet
evaluated on a deeper model or under a different training objective.

\paragraph{Guideline 4 (ICL leads at very low $k$; fine-tuned methods
catch up by $k{=}10$).} Among trained/adapted baselines, prefer
in-context learning at the lowest shot counts ($k{=}2$--$3$). By
$k{=}10$, at least one of SetFit or LoRA has caught up with or overtaken
ICL on every dataset tested. Evidence: Figure~\ref{fig:baselines}.
Caveat: tested only on TREC and AG News (4 and 6 classes); no ICL
comparison was conducted on Banking77 or CLINC150 because of the
class-count constraint discussed in Section~\ref{sec:setup}.

\paragraph{Guideline 5 (E5-Mistral's disadvantage is largely an
estimation-efficiency effect, and partly depends on the classifier).} If
more labeled examples are available, consider increasing $k$ before
concluding that a high-dimensional, instruction-tuned decoder embedding
model is unsuitable. With the tuned probe, its accuracy gap to BGE-large
shrinks by more than 80\% from $k{=}2$ to the largest $k$ tested
($k{=}30$ on Banking77, limited by its smallest class; $k{=}50$ on
CLINC150), and every paired comparison reaches $p{=}0.00195$
(Section~\ref{sec:discussion}). Part of the gap at small $k$ also depends
on the classifier: with shrinkage LDA instead of the probe, the Banking77
gap at $k{=}2$ is $0.117$ rather than $0.217$
(Table~\ref{tab:estefficiency}). Mechanism: not established; these
results show \emph{that} more examples and whitening reduce the gap, but
not \emph{why} this embedding model needs them. Caveat: the larger-$k$
evidence covers the two many-class datasets, and for the regularized
methods only Banking77.

\paragraph{What these guidelines do not cover.} We do not explain
\emph{why} the raw geometry of E5-Mistral is unfavourable for
distance-based rules. We also do not isolate the $n/d$ ratio from
embedding dimension, encoder architecture, and class count, or establish
a general rule for choosing between SetFit and LoRA: we tested and
rejected the hypothesis that class count explains their crossover
(Section~\ref{sec:results:baselines}). These boundaries define the scope
of the findings.

%% file: sections/discussion.tex
\subsection{Three Separability Metrics, Revisited}

Why does E5-Mistral underperform BGE-large and MiniLM in this regime,
despite comparable or better embedding quality by other measures? The
first version tested three geometric metrics of increasing statistical
sophistication and interpreted all three as failures.

\textbf{Raw pairwise cosine similarity}, a measure of anisotropy, does
not produce a consistent ordering across datasets: E5-Mistral has the
highest anisotropy on Banking77 but not on CLINC150. \textbf{The
within/across-class margin} (Section~\ref{sec:results:layers}) is
informative within one model's layers but ranks MiniLM's overall margin
above BGE-large's despite MiniLM's lower accuracy, so it fails across
architectures. Both conclusions still hold. \textbf{A Fisher-LDA-style
whitened separability spectrum}, which reweights directions by inverse
total variance rather than treating every dimension as equally
informative, ranks E5-Mistral as the \emph{most} separable of the three
encoders. This ranking holds across two sample-size regimes and a wide
ridge sweep in each (the full training set,
$\mathrm{reg} \in [10^{-3}, 10]$; and a $k$-shot-sized sample, where
$S_t$ is itself rank-deficient, $\mathrm{reg} \in [1, 50]$). The first
version regarded this ranking as contrary to the accuracy results,
because the probe and nearest centroid rank E5-Mistral last.
Section~\ref{sec:results:whitening} partly changes this interpretation.
Classifiers that whiten with a regularized within-class covariance
recover a large part of E5-Mistral's deficit, so the whitened metric was
measuring a separability that classifiers based on the raw metric do not
use. The metric is not fully supported, however: even with shrinkage
LDA, E5-Mistral reaches the better of BGE-large's two baselines (probe or
nearest centroid) in only 4 of 16 settings, all on TREC, where
E5-Mistral was already ahead. We therefore narrow the first version's
statement that embedding anisotropy is ruled out as an explanation.
\emph{Raw} anisotropy does not predict the accuracy ordering across
datasets, but the raw geometry of E5-Mistral does appear to be
unfavourable for distance-based classifiers.

The remaining part of the deficit is an \emph{estimation-efficiency}
effect. We extended the support set beyond the range otherwise covered
in this study ($k{=}20,30$ on Banking77, limited by its smallest class;
$k{=}20,30,50$ on CLINC150). The tuned-probe accuracy gap to BGE-large
shrinks from $0.19$--$0.22$ at $k{=}2$ to $0.03$--$0.04$ at the largest
$k$ tested on each dataset (Table~\ref{tab:estefficiency}), a reduction
of more than 80\% on both. Every paired comparison (10/10 seeds showing a
narrower gap, on both datasets and for both the tuned probe and nearest
centroid) reaches $p{=}0.00195$, the minimum attainable two-sided
$p$-value with 10 paired seeds. On Banking77, the regularized methods
reach a similarly small remaining gap from a smaller starting gap
($0.117 \to 0.025$ for shrinkage LDA). These results show that, within
the $k{\le}10$ range otherwise reported in this paper, the practical
disadvantage is largely a small-sample effect rather than a permanent
limit. They do not explain why E5-Mistral needs more examples.

\subsection{Why the Encoders Leave Little for Discriminant Structure to Add}

On the two bidirectional encoders, nearest centroid and the tuned probe
are difficult to outperform, and the regularized methods improve on the
original variants without surpassing these baselines. One interpretation
is that pretraining changes the nature of the small-sample problem. In
the classical settings for which KLPCDA was designed, the raw features
carry the class information only weakly, and structure such as $S_b$ and
$S_w$ is what allows a few labeled examples to recover it. A frozen
pretrained embedding has already organized the space semantically, so
class means are often sufficient, and the additional structure imposed by
a discriminant adds less. Whitening helps where this organization is not
suited to a distance-based rule, as on E5-Mistral. This interpretation is
consistent with our results, but we did not design an experiment to test
it.

\subsection{The Degeneracy Beyond Sentence Embeddings}

As independent evidence that the degeneracy itself
(Section~\ref{sec:method:degeneracy}) arises from the construction of
the estimator rather than from the frozen sentence embeddings studied in
this paper, we separately verified it on a structurally different type of
data: residual-stream activations extracted from a decoder-only
generative language model during generation, rather than fixed output
embeddings. The same exact rank-$(n{-}L)$ and eigenvalue-collapse result
held to machine precision, and the tie-break
(Section~\ref{sec:method:fix}) again recovered a direction that was
stable across independent resamples, unlike the untied basis. We do not
pursue this domain further in this paper; we report it only as evidence
that the mechanism generalizes beyond frozen sentence embeddings.

\subsection{Limitations}

\textbf{The tie-break direction remains unexplained.} In the truncated
implementation, ranking tied directions by ascending variance improves
accuracy, whereas the literal ``maximize'' reading reduces it
(Section~\ref{sec:method:fix}). In the regularized family, the leading
$L-1$ directions need no tie-break, but the remaining block is split at
first order in $\mu$ by within-class variance, apparently in the opposite
(descending) order \citep{qu2026modular}. Whether the two criteria
measure the same quantity, and why ascending order improves accuracy,
remain open questions.

\textbf{Choices in the regularized protocol.} At $k=2$, every inner
cross-validation fold has one example per class, so $S_w$ vanishes, all
values of $\mu$ give the same score, and the ridge defaults to the middle
of the grid. The projection is followed by 1-nearest-neighbor, the
original convention, rather than by a classifier tuned on the projected
features, and the subspace dimension follows the original fixed rule;
the main result depends on that rule (Section~\ref{sec:results:headline}).
The larger-$k$ test of the E5-Mistral effect covers Banking77 only.

\textbf{Baseline dataset coverage is not fully unified.} SetFit and LoRA
were evaluated on all four datasets, whereas in-context learning was
evaluated only on TREC and AG News. Including demonstrations for 77--150
classes in a single prompt is impractical because of context-length
constraints and per-example generation costs. Supporting ICL at higher
class counts would require a different design (e.g., retrieval-based
candidate-label shortlisting), which would not be directly comparable to
the exhaustive-listing ICL used here.

\textbf{Other limitations.} We use a linear kernel throughout, matching
the linear and polynomial kernels commonly reported in the source papers'
experiments, but leave RBF and other kernel choices untested on
embeddings. All experiments use 10 random $k$-shot draws per condition; a
larger number of seeds would make every comparison more precise. Several
design choices, including the datasets, embedding sizes, number of seeds,
and test-set subsampling for in-context learning on AG News, were
constrained by the available compute on a single GPU machine. We document
these choices rather than presenting the study as more exhaustive than
it is.

%% file: sections/conclusion.tex
We asked what small-sample discriminant analysis contributes to few-shot
text classification on frozen LLM embeddings, and which of its components
is responsible. The answer has five parts. First, the standard balanced
$k$-shot protocol makes the truncated implementation of KLPCDA exactly
degenerate, in closed form, and makes its results depend on the
floating-point environment. Second, replacing the truncation with a ridge
selected within the support set removes both problems and outperforms
the best original variant in 45 of 48 settings, so choosing among the
original variants is not worthwhile. Third, in general the regularized
formulation does not outperform a tuned probe or nearest centroid on the
bidirectional encoders we tested. Fourth, where it does help, on the
decoder-based E5-Mistral, the component responsible is regularized
within-class whitening, which a standard shrinkage LDA provides equally
well. Fifth, most of the remaining deficit of E5-Mistral depends on how
many labeled examples are available. Two questions remain open: whether
the whitening effect holds for other decoder-based embedding models, and
why ascending variance is the better tie-break in the truncated
implementation. We release the estimator, the evaluation code, and the
per-seed results.